# Assessing the impact of regulations and standards on innovation in the field of AI

**Alessio Tartaro**
University of Sassari
Italy
a.tartaro@phd.uniss.it

**Adam Leon Smith**
Dragonfly, Barcelona, Catalonia, Spain
Chair, Fellows Technical Advisory Group, British Computer Society
adam@wearedragonfly.co

**Patricia Shaw**
Beyond Reach Consulting, United Kingdom
Trustee Director and Chair, Society for Computers and Law
Director, ITechLaw Association
trish@beyondreach.uk.com

## Abstract

Regulations and standards in the field of artificial intelligence (AI) are necessary to minimise risks and maximise benefits, yet some argue that they stifle innovation. This paper critically examines the idea that regulation stifles innovation in the field of AI. Current trends in AI regulation, particularly the proposed European AI Act and the standards supporting its implementation, are discussed. Arguments in support of the idea that regulation stifles innovation are analysed and criticised, and an alternative point of view is offered, showing how regulation and standards can foster innovation in the field of AI.

## 1 Introduction

Over the past few years, AI has grown at an unprecedented pace thanks to the increasing availability of data, the development of sophisticated machine learning techniques, and the surge of computational power. This has led to the ubiquitous application of AI in several areas, including employment, education, access to essential private and public services and benefits, law enforcement, administration of justice and democratic processes. While the use of AI in these areas is bringing clear benefits, it is also creating new risks. Many cases have already been reported where AI has created discrimination, threats to people's safety, health, and fundamental rights, and undermined the stability of democratic institutions. To minimise these risks and maximise the benefits, a new wave of AI laws is emerging, notably the proposed European AI Act, to regulate the use of high-risk AI systems.

Although regulations seem prima facie necessary to avoid the negative consequences of unconstrained uses of AI, some have criticised this approach, arguing that regulations will stifle innovation. This position is supported by numerous arguments to prove the opposite thesis, namely that innovation can flourish only with little or no regulation. Regulations are thus seen as an obstacle to the free development of market-driven innovation that would proceed seamlessly in the absence of regulatory barriers.

This paper critically examines the idea that regulations will stifle innovation in the field of AI. First, we examine current trends in the regulation of AI, focusing on the proposed AI Act and the role that standards play in demonstrating compliance with the requirements and obligations of the proposed Regulation. Next, we provide a series of arguments supporting the idea that regulations (and consequently standards) will stifle innovation. Then we analyse and criticise these arguments. Finally, we offer an alternative point of view, showing how regulations and standards can foster innovation.

## 2   The proposed AI Act and the role of standards

The proposed European Union (EU) AI Act [9] aims to take up the opportunities offered by AI while also respecting safe and ethical boundaries and ensuring the development of trustworthy AI in the EU. The proposed AI Act adopts a risk-based approach, dividing AI applications into four categories: unacceptable risk, high risk, limited risk, and minimal risk. Unacceptable risk AI systems, such as those used for social scoring, are proposed to be banned under the new Regulation due to the potential harm they could cause to the health, safety, and fundamental rights of individuals. High-risk AI systems include those used, among others, in education and vocational training, employment, workers management and access to self-employment, law enforcement, administration of justice and democratic processes. High-risk AI systems are subject to specific requirements covering the following areas: risk management, data quality and governance, technical documentation, record keeping, transparency and provision of information to users, human oversight, accuracy, robustness and cybersecurity. Limited risk AI systems are subject to specific transparency obligations which serve to inform users about their interaction with AI systems. Finally, minimal or no risk AI systems, which include most AI systems developed and provided in Europe, are not subject to any requirements or obligations.

The proposed AI Act builds on the New Legislative Framework (NLF) [4]. The NLF was introduced in the mid-1980's and involves the delegation of technical regulations to standardization bodies through harmonized standards. Under the NLF, the legislation defines only essential requirements for products and leaves the technical specification of these requirements to the European harmonized standards. Compliance with the legislation is then assessed against these harmonized standards.

The creation of a set of technical specifications is delegated to European Standardisation Organisations such as CEN/CENELEC. National authorities are required to recognise that products manufactured according to the standards of these private organizations are presumed to conform with the essential requirements specified in EU Directives or Regulations. The NLF is implemented through a mix of market surveillance, third-party conformity assessment, and self-assessment. Compliance with such standards is voluntary in theory, however the burden of proving that alternative approaches meet the obligations of the law is potentially very costly [27].

Many standards adopted in Europe are directly adopted International Standards. ISO/IEC is the primarily International Standards settings body, and it also came to prominence in the 1980s. This was driven by the Agreement on Technical Barriers to Trade that requires members of the World Trade Organisation to use International Standards as the technical basis of domestic laws and regulations - unless they are ineffective or inappropriate [29].

The same regulatory approach can be found in the proposed AI Act. The proposed Regulation includes provisions for the use of harmonized standards to assist providers of high-risk AI systems in complying with the requirements of the Regulation. The use of standards is intended to facilitate understanding of the regulation and to minimise the costs of compliance. Article 40 provides that high-risk AI systems which are in conformity with harmonized standards or parts thereof should be presumed to be in conformity with the requirements of the Regulation. Accordingly, providers of high-risk AI systems may demonstrate compliance with the Regulation by complying with officially adopted harmonized standards that cover the requirements of the Regulation. The European Standards Organizations are responsible for preparing these standards, following a standardisation request from the Commission. Once the standards are complete, they are evaluated by the Commission and, if approved, published in the Official Journal of the EU. Only standards published in the Official Journal give AI providers the assurance that their products meet the legal requirements of the proposed AI Act [28].

The Commission recently sent a draft standardisation request to CEN/CENELEC [2]. The proposal lists ten areas in which the two European Standards Organisations are called upon to draft and adopt standards. These standards will serve as the basis for the development of harmonized standards to support the proposed AI Act. Although the debate on the appropriateness and effectiveness of this approach is still ongoing [34], the proposed AI Act represents a milestone in the regulation of AI.

Nevertheless, a comprehensive analysis of its impact on innovation has not yet been conducted. While some argue that regulation will stifle innovation, we argue that regulation accompanied by clear requirements encoded in harmonized standards can support and foster innovation, especially by small and medium size enterprises (SMEs), and create a level playing field. We address this issue in the remainder of this paper.

## 3   Arguments for *regulation stifles innovation*

The idea that regulation stifles innovation is not new [30]. In the field of AI several arguments have been advanced to show how regulation will stifle innovation. In this section, we analyse some of them.

Table 1: Summary of arguments in [18]

| | |
|---|---|
| Slower and more expensive AI development | Regulations make the development of AI slower and more expensive because they require firms to establish processes to comply with the requirements. Delays due to conformity can lead to a loss of competitiveness, while costs can be a barrier to entry for SMEs and start-ups. |
| Less innovation | Bans on certain applications of AI will prevent reaping its benefits. For example, a possible ban on autonomous vehicles would lead to the loss of the benefits that would result if they were adopted. |
| Lower-quality AI | Requirements on explainability may lead to the development of lower quality AI systems due to the trade-off between accuracy and explainability. |
| Less AI adoption | Requirements on human oversight may reduce the adoption of AI. Requiring a human to review the decisions of an AI system leads to a loss of the benefits (speed, accuracy) that result from adoption. |
| Less economic growth | Regulations based on the precautionary principle discourage AI adoption and hinder economic growth because they slow down the increase in productivity. |
| Fewer options for consumers | When certain AI systems are banned, consumers lose the ability to access products or services based on those applications. |
| Higher prices | Regulations create compliance costs and these are passed on to the consumer. Furthermore, the ban on certain applications prevents the costs of certain processes from being reduced. |
| Inferior consumer experiences | The ban on certain AI applications such as facial recognition prevents the provision of better services for consumers, e.g., supermarkets without cashiers. |
| Fewer positive social impacts | The ban on some applications prevents its positive social effects from being realised. For example, autonomous vehicles can use platooning to reduce emissions and bring environmental benefits. |
| Reduced economic competitiveness and national security | Finally, the regulation of AI damages the countries where it is adopted, because it gives an economic and national security advantage to countries where AI is not regulated. |

The piece by Daniel Castro and Michael McLaughlin of the Information Technology and Innovation Foundation (ITIF) [18] offers the most comprehensive list of arguments in support of the idea that regulation will stifle innovation in AI. According to the authors, the regulation of new technologies can be guided by two alternative principles: the precautionary principle and the innovation principle. The meaning of the precautionary principle and the ways of applying it have been the subject of debate since it was first formulated in the 1970s [5]. The authors adopt the strongest and most stringent interpretation, according to which the precautionary principle requires that the use and adoption of new technologies should be restricted or prohibited until they are proven to be safe. In contrast, the innovation principle assumes that most technologies are beneficial, and that they should only be restricted when proven to be dangerous. The two principles result in diametrically opposed legislative choices. On the one hand, legislation based on the precautionary principle places limits on the use of a new technology. On the other hand, legislation based on the innovation principle adopts a case-by-case approach, in which limitations are only imposed when dangerous applications emerge. For example, internet regulation in the US was guided by the innovation principle, while regulation related to genetically modified organisms in Europe is guided by the precautionary principle. Although the authors never mention the AI Act directly but generally speak of 'AI policies', it is clear that the proposed Regulation is a piece of regulation based on the precautionary principle, although it does not adopt the strictest interpretation of the principle. The AI Act does not restrict or prohibit the development and use of AI (with very few exceptions), but it places strict conditions when AI systems pose high-risks. According to the Castro and McLaughlin, this type of regulation harms innovation in ten ways, listed and commented on in 1

Many of these arguments do not directly concern the effects of regulations on innovation. Rather, they concern other effects of regulations, for example, on consumers, society, economic growth and competitiveness, and national security. Although these effects are indirectly related to innovation, they are not considered here.

If particular applications of AI are banned, it is for a reason relating to fundamental rights. Additionally, the argument about quality is a misunderstanding of regulators' position on this — no regulator has proposed requirements for local explainability that may degrade accuracy. Regarding the direct impact of regulation on innovation, the authors' position can be summarised as follows: regulation stifles innovation in the field of AI because it makes the development of AI systems slower and more expensive, decreases their quality, creates barriers to entry for smaller players, and discourages the adoption of innovative AI applications. In section 3.2., we analyse this topic and highlight some of its weaknesses.

Another study analyses more specific challenges and consequences of the regulation of AI in the field of AI-based healthcare technologies [25]. Through interviews with Finnish healthcare stakeholders, the authors identify four main issues that may have a negative impact on innovation in this field. First, ambiguous regulations and divergent interpretations by different authorities can hinder innovation. Second, the complexity of regulations can be an obstacle to their proper implementation. Third, uncertainty on liability issues may slow down the adoption of AI solutions. Finally, limitations on the use of personal data may limit the development of AI products that, in the medical sector, need to use personal data sometimes without the user's explicit consent.

The same study also investigates possible consequences of AI regulation. According to the stakeholders interviewed, strict regulations will lead to a loss of competitiveness and the relocation of business players to countries and regions where they are less regulated. Furthermore, the overlapping of different regulations in the medical sector could lead to an over-regulation of some AI-based healthcare technologies. For instance, wearable devices could be regulated by both the Medical Devices Directive and the proposed AI Act. All this could lead to a deceleration and a barrier to the development of these technologies, but also to an increased demand for professional services dealing with regulatory issues.

## 4 A more nuanced view on the relationship between regulations and innovation

The arguments in the previous section raise issues that need to be taken into account. However, they offer a partial perspective on the relationship between regulation and innovation. First, they have a limited view on what innovation is and on which actors have a say on its developments. It is significant, for example, that in Konttila and Väyrynen's study [25] neither patients nor doctors are included among the interviewed stakeholders, while entrepreneurs and lobbyists seem to be overrepresented. In addition, these studies neglect how regulations can have a positive effect on innovation. While Castro and McLaughlin [18] argue that the innovation principle is good and the precautionary principle is bad when regulating new technologies, they overlook numerous cases where regulations based on the precautionary principle stimulate innovation. For instance, environmental regulations had a positive effect on green innovation [7], and data protection principles have boosted the development of privacy-enhancing technologies [17]. These critical remarks suggest that the relationship between regulation and innovation is more complex than previously claimed. Empirical and theoretical studies on the relationship between regulations and innovation show that, in different scenarios and under different conditions, regulations can either encourage, stifle, or have no effect on innovation.

### 4.1 How (global) regulatory vacuum impacts innovation

The impact assessment accompanying the proposed AI Act describes several ways in which the proposed regulation may have an impact on innovation in the field of AI [7]. The impact assessment notes how the current regulatory vacuum can have a negative impact on innovation in three ways. First, legal uncertainty is a major obstacle for investments in AI. Without clear rules on requirements and obligations for AI systems, on the distribution of responsibilities, and on certification and assurance mechanisms, the uptake of AI is unlikely to be successful. Second, the distribution of unregulated AI systems on the market decreases the safety of these products and increases the risk of their negative impact. This results in a lack of consumer trust and reduced social acceptance. Private and business consumers' mistrust represents an obstacle for investments and for the adoption of AI. Finally, the lack of common European legislation can lead to a fragmentation of the internal market, which can have a negative impact on innovation in many respects (e.g., difficulties in accessing national markets especially for SMEs and start-ups, shift of investments to countries with light regulation, lack of interoperability, etc.).

We live in an increasingly global world, and the nature of the AI ecosystem demonstrates this most clearly. AI operators and users can innovate from anywhere within the global market, such that the multiple geographical locations of where an AI system is designed, developed, and deployed can make the implementation of AI borderless. A lack of international regulatory cohesion, can itself cause legal uncertainty and be a barrier to entry for those who operate in multiple jurisdictions, and can create instability in global AI competences. This global fragmentation can undermine the safety of AI and the trust that can be placed by end users in the governance of organisations and the quality of the AI created. The lack of consistency leaves both businesses and consumers exposed. This can be seen through the plethora of Data Protection and Privacy Regulator interventions required in recent years across the globe concerning Clearview AI [10]. Furthermore, the lack of interoperability of laws between nations and economic blocks (such as the EU) may not directly limit innovation but can make the cost of bringing innovation to all in the global market more costly for some jurisdictions than others.

This disparity between jurisdictions who deploy AI regulations and those do not, could result in certain jurisdictions being seen as AI "safehavens" or justify the perceived "benefits" of unregulated marketplaces. At best, this may cause law forum-shopping for AI firms, but at worst it would lend itself to making societally acceptable the fact that some countries and people groups will be put at a disadvantage (put at a safety risk or have more risky and less trustworthy AI used with them) over others. Depending on the AI and data literacy of the jurisdiction, unscrupulous international business may even legitimise their exploitation of and harvesting of training data[1] from such countries and people groups, all in efforts to avoid regulation.

Whether the EU's proposed AI Act has the "Brussels effect" [31] or not, in recognition of the global economy and the nature of the global AI ecosystem, there is and will be a clear need for coherence in and a systematic approach to global AI regulation [26]. In contrast, standardisation, particularly standards which are adopted by global standards bodies such as the ISO and IEEE, can facilitate technical interoperability and consistency of approach to AI safety and governance through both technical and socio-technical standards. This has been demonstrated through harmonisation of wireless protocols [11]. These International Standards can align both safe, secure, trustworthy operations (through methods, processes, recommended practices, procedures) with the quality of the output and with globally beneficial and fundamental rights preserving outcomes. By so doing, standardisation can mitigate the negative effects of global fragmentation and help bridge the gap between inconsistent regulatory regimes.

Depending on where an AI system is being designed, developed and deployed and how much levelling up of the AI system needs to be done to meet the standards, can mean that (at least initially) AI projects may be more expensive in the short term. In the longer term, more markets in the global economy are being opened to the supplier of the AI system, as well as producing a better end product, which over time should result in those upfront costs being amortised. Standards can themselves be innovation enhancing as they enable businesses to build upon the best or good practice of others in the industry, and can increase the probability of market up-take of technological innovations [6].

Compliance with standards, can provide businesses with certainty that other operators within their AI supply chain have achieved a certain level of safety, quality, and trustworthiness. This is particularly important given that many obligations in the proposed EU AI Act are on the procurers of AI systems, not the producers. Furthermore, businesses engaging in International Standards making can have a positive influence on innovation [15] as well as providing certainty for them operating in a global market. Standards (if accompanied with an independently verified internationally recognised certification or Trustmark) can also win the trust of and advance adoption by customers. Socio-technical standards in the field of AI are increasingly requiring those who adopt them to be more transparent and engage with their stakeholders [3], both of which facilitate an important feedback loop between customer and supplier to identify and help resolve complaints, pain points and friction, and to become aware of more innovative approaches to fulfilling the needs of a business' customer base.

For standards to work most effectively, they need to be adopted and recognised by industry, usually voluntarily, but quasi-mandated through a series of contractual obligations across the supply chain or between industry bodies and their members. Whilst mandating International Standards in law would ensuring regulatory compliance by all in the AI ecosystem, it is often through the standards reaching uncodified "best practice" status, which if not adhered to means that industry players are penalised by their customers, industry bodies and/or sectoral regulators, where International Standards can maintain quasi-mandatory status.

Certainty can be provided by standards that mitigate the impacts of uncertainty and make innovation more accessible in a global legal and AI landscapes by bridging the gaps where individual jurisdictions or economic blocks are powerless to do so.

When the proposed AI Act is agreed to close this regulatory gap, its impact on innovation will depend on two contrasting factors. On the one hand, the proposed AI Act can have a negative impact because it makes AI projects more expensive due to compliance costs and administrative burdens. On the other hand, the regulation can promote uptake through clear rules, legal certainty, higher trust, and greater social acceptance. The proposed AI Act can be successful if it manages to ensure a good balance between these two factors. In this regard, the impact assessment argues that the proposed AI Act approach, i.e., mandatory requirements for high-risk AI applications combined with voluntary codes of conduct for non-high-risk AI applications, achieves this balance. In fact, only high-risk AI systems, which are a small percentage of all developed AI systems[2], will undergo additional compliance cost, while the others will benefit from the positive impacts of the Regulation. Furthermore, although more expensive, high-risk systems will also enjoy greater trust and acceptance that will favour their adoption in the European market.

As mentioned previously, standards play a crucial role in implementing the requirements of the proposed AI Act. For this reason, it is important to analyse how standards, in addition to legislation, can have an impact on innovation. An analysis of the impact of AI standards on innovation in the field of AI is still lacking. However, some studies have analysed the relationship between standards and innovation in other fields and on a general level. These studies show that standards, like regulation, can have a variable effect on innovation.

---

[1]Training data is a subset of input data samples used to train a machine learning model.
[2]Estimates range between 5% and 33% [12]

## 4.2 Regulations, standards, and innovation. Conceptual framework

Standards have four main economic functions, which can have positive or negative effects on innovation depending on the circumstances [17] [14]. First, standards codify knowledge by setting out information that can be used to innovate on top of complex standardised technology. Second, standards reduce the variety of options that allow for economies of scale and critical mass in emerging areas, as well as promote cohesion in a sector and encourage market growth. However, reducing variety may also have negative effects: it encourages incremental innovation instead of radical innovation, may produce market concentration, and may lead to a premature choice of the dominant technological solution. Third, standards define minimum levels of quality to build trust among early adopters of emerging technologies and avoid incidents that could damage trust in new products. Quality standards can also help to prevent low-quality goods from driving out high-quality products in markets where there is a high level of information asymmetry (known as Gresham's Law). Finally, standards increase interoperability that fosters market integration. However, this can also create lock-in effects and monopolies.

A study by Blind, Petersen and Riillo investigates under which conditions regulations and standards can have positive or negative effects on innovation [15]. Their analysis focuses on three concepts, i.e., market uncertainty, regulatory capture, and information asymmetry. The authors consider legislations and standards as two alternative forms of regulation, the former being government-led and mandatory, the latter being industry-led and voluntary.

Market uncertainty refers to the lack of clarity about the state of a market caused by factors such as competition, consumer behaviour, and technological complexity. Market uncertainty can be particularly high in areas where there is no dominant technical solution, resulting in unpredictable behaviour of consumers, who are unable to assess the quality of different options and make informed purchasing decisions, or comply with applicable regulation. Producers may also have difficulty predicting the direction of technological development in such markets. In their study, the authors consider technological uncertainty as the main factor contributing to market uncertainty. Given the rapid developments in the field of AI and the lack of a dominating technical solution, the AI market can be considered an uncertain market under this definition.

Regulatory capture refers to attempts by interest groups, such as industries, to influence government regulations in a way that favours their own interests. Both government regulations and International Standards are vulnerable to regulatory capture, with the latter more exposed than the former due to the openness of standardisation processes. CEN/CENELEC, for example, has been criticised for a lack of diverse stakeholder participation [34] and the European Telecommunications Standards Institute (ETSI) was excluded from the European Commission's draft standardisation request on the AI Act because it was considered too private-sector driven [13].

The impact of regulatory capture on formal standardisation may vary depending on the level of uncertainty in the market, with firms being more able to influence standards to align with their preferences in markets with low uncertainty, but having less success in highly uncertain markets. Influencing the standardisation process can give a company a competitive advantage, favouring the adoption of its own technology preference and thus raising the costs of other companies adopting a different technology solution. An example of this kind of regulatory capture practices in standardisation activities was Blake Lemoine, fired from Google, was sent to ISO/IEC AI standardisation meetings to pursue Google's agenda and ensure standards had a very narrow or very wide definition of AI [16].

In the context of technological innovation and regulation, information asymmetry refers to the difference in knowledge between regulators and market players about the state of the art in a given technology sector. In cases where regulators have less knowledge of the technological frontier than market actors, this information asymmetry can lead to a discrepancy between existing regulations or standards and the actual opportunities offered by the technological frontier. This can have a negative impact on the innovation processes of companies, which experience regulations and standards as an obstacle. Information asymmetry can also increase market uncertainty, which can further exacerbate the potential for technological misalignment and increase the innovation costs associated with regulatory compliance.

Based on these conceptual considerations and analysing empirical evidence, the authors reach the following conclusions on the effect of regulations and standards on innovation depending on the degree of market uncertainty. In highly uncertain markets, regulations are shown to have higher compliance and consequently innovation costs than standards due to higher levels of information asymmetry, while the effects of regulatory capture are similar for both regulations and standards. In low uncertain markets, standards are linked to higher compliance and innovation costs than governmental regulation because they are more prone to regulatory capture, while the effects of information asymmetry do not differ between regulations and standards. In other words, standards may benefit innovation more than governmental regulations in highly uncertain markets, while in low uncertain markets regulations may have better effects than standards.

These results are not directly applicable to the case of regulation and standardisation of AI. In the study, in fact, regulations and standards are considered as distinct and alternative regulatory instruments. The former are mandatory rules issued and enforced by the government, while the latter are created by recognised standardisation organisations and are voluntary and based on consensus. However, the authors do not consider cases where regulations and standards are complementary rather than alternative. This is what happens with legislation under the NLF, including the proposed AI Act. For this reason, it is necessary to adapt and expand these results to the case of the complementary regulation and standardisation in the field of AI. We discuss this topic in the next section.

## 5 Discussion

The studies analysed in the previous section suggest that the impact of regulation and standards on innovation in the field of AI will depend on the level of uncertainty in the AI market. Given the rapid developments and lack of a dominant technical solution in the field of AI, the AI market can be considered a highly uncertain market. Accordingly, Blind's studies suggests that standards may benefit innovation more than governmental regulations due to higher levels of information asymmetry and similar effects of regulatory capture. Standards, being industry-led and voluntary, may provide more flexibility and clarity for companies to navigate the uncertain market and align with the technological frontier.

This conclusion is supported by the high level of information asymmetry between regulators and market players (and also among market players) in the field of AI. Since regulators have less knowledge of the technological frontier than market actors, this information asymmetry can lead to a discrepancy between proposed regulations and the actual opportunities offered by the technological frontier. This can have a negative impact on the innovation processes of companies, which experience regulations as an obstacle. This is a further reason to support the use of industry-led standards for the regulation of AI. When regulations and standards align with the technological frontier, they do not impede innovation. Similarly, when the technological frontier proceeds within the boundaries set by regulations, standards represent an opportunity to foster innovation in the field of AI for the benefit of all stakeholders.

### 5.1 How regulation and standards can foster innovation in the field of AI

In order to extend this analysis, it is useful to compare the draft EU AI regulation to the GDPR [22]. The GDPR is not supported by standards, however it does allow for third-party GDPR certifications to be awarded. These certification schemes do not provide presumption of conformity with GDPR [1]. Such certification schemes have also been slow to manifest, and there is only one that is endorsed by the European Data Privacy Board [20].

The complaint-based and enforcement-led approach of GDPR means that levels of compliance are determined by each organisation's judgement of the risk and impact of fines and reputational damage, and limited by the number of resources available to data protection authorities. Additionally, there is growing evidence that GDPR has resulted in less competition in some areas, and favoured large online platforms [23]. One of the reasons for this is that companies are less willing to trust smaller companies with data, as it may impact their own compliance [24]. Consequently, the decision to use regulations without standards has had a substantial impact on innovation in the uncertain data sector. It should be acknowledged, however, that this impact has not only been negative. The emergence of privacy-enhancing technologies is linked to the GDPR, and represents a positive impact of the regulation on innovation.

Many innovative developments come from small enterprises, who have some advantages over larger companies [35] [8] [21]. However, one disadvantage that startups usually have is a lack of financial resources. As Forbes magazine pointed out [32], the legislation is deliberately vague — much like the EU's proposed AI act it is vague in order to remain future-proof. The result of this is that many large firms are spending as much of 40% of their GDPR compliance budget on legal advice in order to interpret the meaning of many of the requirements. Clearly, smaller firms are less able to rely on legal advice, and less able to pay fines if the legal advice is not supported by decisions by the regulators.

In contrast, the proposed AI Act is based on conformity assessment processes prior to placing a product on the market. Conformity assessments require clear and unambiguous technical standards. Such standards are obtainable by businesses for a negligible fee. Conformity assessments come at a cost, but not a significant cost compared with the cost of actually complying with the requirements [19], or obtaining expensive legal advice.

Innovative businesses that remain unclear about compliance can participate in regulatory sandboxes, expected within each member state, that allow a "safe space" to work with regulators on innovation. Since the UK Financial Conduct Authority launched its sandbox in 2016, it has supported more than 700 firms and increased their average speed to market by 40% compared with the regulator's standard authorisation time [33]. However, sandboxes are limited by a number of factors. Firstly, the willingness of companies to subject themselves to regulatory scrutiny before they need to. Secondly, the ability of companies to share data with regulators. While the proposed EU AI Act modifies the GDPR to allow companies to process data in order to comply with aspects of the regulation, the sandboxes are expected to be in operation long before the proposed AI Act.

Also, as previously noted, regulation in some areas has sparked innovation. It is likely similar innovations will result from AI regulation, this may be in technical compliance and post-market monitoring, and may be in the creation of new kinds of roles and professions to help govern AI systems. Early work by standardisation bodies to define the criteria for competent persons to assess AI systems indicates that no single person is likely to have all the skills required - as this encompasses a diverse set of socio-legal-technical competencies. Technical resources are unaccustomed with considering legal and ethical requirements, legal experts rarely have data science skills, and AI ethicists are more focused on the broad and long-term impacts of AI systems on individuals and society. Therefore, a comprehensive assessment of AI systems requires a team of experts from different disciplines, including technologists, legal and ethical experts. The development of this multi-disciplinary expertise and the establishment of new processes to combine and integrate them within organisations might be one of the major innovations brought about by the regulation of AI.

# 6 Conclusion

In summary, innovation will be stifled by regulation in areas where AI applications are prohibited. However, in areas where they are considered high-risk, innovation will be supported by clear technical requirements, and the opportunity to directly discuss and test ideas with immediate regulatory feedback. Innovators will not need legal advice to comply with clear standards. Conformity assessment processes will mean procurers of AI systems and services will be able to rely on recognised certifications to ensure they can have confidence in buying from smaller, more innovative companies.

However, it is also noted that there are over 40 standards so far that could be in consideration to support the proposed EU AI Act, and until they are finalised, it will not be clear to businesses how they can comply. If harmonized standards are not ready when the AI act takes effect, then there will be a period of time when innovation will be stifled, because only companies with significant legal or compliance teams and capital available to pay fines will be able to innovate.